\documentclass[a4paper, 10pt, conference]{ieeeconf}      
\IEEEoverridecommandlockouts                              

\usepackage{graphicx} 
\usepackage{epsfig} 
\usepackage{mathptmx} 
\usepackage{times} 
\usepackage{amsmath} 
\usepackage{amssymb}  
\usepackage{tabularx}
\usepackage{subcaption}
\usepackage{fancyhdr}
\fancypagestyle{withfooter}{
  
  \fancyfoot[C]{\footnotesize Accepted to the IEEE IROS workshop on Autonomous Robotic Systems in Aquaculture: Research Challenges and Industry Needs}
}

\title{\LARGE \bf
Biology and Technology Interaction: Study identifying the impact of robotic systems on fish behaviour change in industrial scale fish farms
}

\author{ 
L. Evjemo$^{1,\dagger}$, Q. Zhang$^2$, H. G. Alvheim$^{2}$, H. B. Amundsen$^{1,2}$, M. Føre$^{2}$, E. Kelasidi$^1$  \\ 
\thanks{$^1$ SINTEF Ocean AS, Dept. of Aquaculture Technology, 7010 Trondheim, Norway.}
\thanks{$^2$ Norwegian University of Science and Technology Dept. Engineering Cybernetics, 7034 Trondheim, Norway.}
\thanks{$^\dagger$ Corresponding author: linn.evjemo@sintef.no}%
}

\begin{document}

\maketitle
\thispagestyle{withfooter}
\pagestyle{withfooter}

\begin{abstract}

The significant growth in the aquaculture industry over the last few decades encourages new technological and robotic solutions to help improve the efficiency and safety of production. In sea-based farming of Atlantic salmon in Norway, Unmanned Underwater Vehicles (UUVs) are already being used for inspection tasks. While new methods, systems and concepts for sub-sea operations are continuously being developed, these systems generally does not take into account how their presence might impact the fish. This abstract presents an experimental study on how underwater robotic operations at fish farms in Norway can affect farmed Atlantic salmon, and how the fish behaviour changes when exposed to the robot. The abstract provides an overview of the case study, the methods of analysis, and some preliminary results.

\end{abstract}

\section{INTRODUCTION}

Sea-based aquaculture production has had a rapid growth over the last decades, and the Food and Agriculture Organization of the United Nations (FAO) estimates that this growth will continue~\cite{FAO2024}. Norway is the world's second-largest exporter of aquatic animal products, with a share of 8 \% of the global value of aquatic animal product exports in 2022. Over 73 \% of Norway's aquatic animal export revenue is salmon products, making Norway the largest producer of Atlantic salmon (\textit{salmo salar}) in the world. To meet the growing demand for food from aquaculture, the industry is targeting innovations that can improve the efficiency and safety operations. In Norway there is also an increasing shortage of new salmon production sites near shore, and future developments are expected to push production further out to sea, to more exposed and demanding sites. New solutions should there for increase safety of work both for human operators, for the structures, and for the fish. Some robotic solutions has already been implemented to parts of daily operations at the fish farms, for example net inspections by Remotely Operated Vehicles (ROVs), checking for holes and weaknesses in the net that might lead to escapes. 

\begin{figure*}[htbp]
    \centering
    \includegraphics[width=0.8\textwidth]{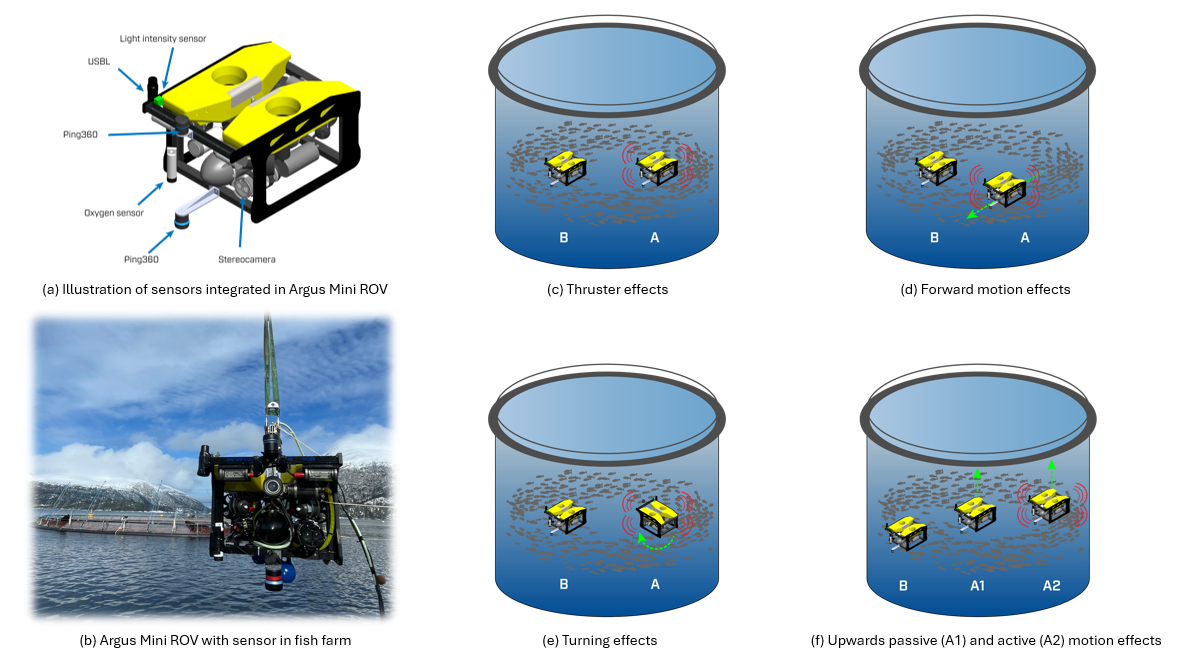}
    \caption{Illustrations of experimental case studies. (a) integrated sensors on ROV; (b) Argus Mini ROV with integrated sensors during deployment at ACE facilities;(c) Stationary position with A: Active thrusters, B: No thruster activity; (d) A: Forwards movement, B: Stationary; (e) A: Turning, B: Stationary; (f) A1: Upwards passive movement, A2: Upwards active movement, B: Stationary}
    \label{fig:illustraioncases}
\end{figure*}

The robotics community continuously develop new systems, concepts and methods for UUVs in sub-sea operations, and such vehicles are used in a wide range of industries from oil and gas to archaeology. However, most of these methods and robotic systems are intended for environments with static and rigid structures. Robot systems working within a net pen at a salmon farm has some extra challenges: The net pen is itself flexible and deformable~\cite{Su2021}, moving with waves and currents close to the surface. Also, the robot needs to interact with and consider the unpredictable movements of 200 000 fish, not hurting them or introducing unnecessary stress or discomfort. There is therefore a need for more research into biology and technology interaction related to fish farming, where preliminary research studies have been conducted to investigate how robotic systems migh impact different species, though mostly in controlled lab settings. 
Farmed fish are exposed to a series of factors that can affect the behaviour, such as changes in lights sources, sounds, intrusive objects in their immediate proximity, and conditions in the water such as temperature, salinity and oxygen levels. To the best of the authors knowledge, not studies have so far attempted to quantify the changes in fish behaviour due to the presence and operation of UUVs suitable for autonomous operations in sea-based aquaculture. This is the focus of the experiments presented in this abstract, showing preliminary results from analyses made on fish behaviour changes during a set of case studies on different ROV operations and movements. An ROV is operated in an industry-scale, fully operational fish farm, and a series of case studies are performed to test the changes in behaviour from the fish. To collect the data, sonars and a stereo cameras were used, mounted on the ROV as seen in Figure~\ref{fig:illustraioncases}b. Preliminary results are presented in this abstract. 

\section{Experimental setup and data acquisition}

The experimental case studies presented in this abstract focused on how farmed Atlantic salmon would react and interact with a ROV of typical size and make of those already used in production of farmed Atlantic salmon in Norway. Field trials were conducted at SINTEF ACE Korsneset, an industrial scale fish farm in central Norway that is part of the SINTEF ACE research facility suite~\cite{ACE}. The site where these experiments were conducted features 15 cages in total, each containing up to 200 000 farmed salmon, or 1000 t of fish. The specific cages used for these trials, cage 13 and 15, each contained approx. 170 000 salmon of mean weight about 3.5 kg. During trials, feeding was stopped, as feeding will strongly impact the behaviour of the fish, breaking away from the normal schooling behaviour~\cite{ferno1995vertical}.

\subsection{Experimental Case Studies}
The Argus Mini used in these field trials is a 90kg observation class ROV that measures $L\times W \times H = 0.9$ m $\times 0.65$ m $\times 0.5$ m. The ROV has four horizontal and two vertical electric thrusters arranged such that the vehicle is actuated in surge, sway, heave, and yaw \cite{fossen_handbook_2011}. As illustrated in Fig.~\ref{fig:illustraioncases}, different movements were tested using the Argus ROV: Thrusters on and off, forwards movements, upwards movements (with and without active thrusters), and turning movements imitating an inspection operation.

\subsubsection{Thruster effects} Testing how the fish reacts to the ROV in a stationary position with and without thruster activity. To enable tests where the ROV did not use thruster power, extra weight was added to compensate for the initial positive buoyancy. While the extra weight was attached to the ROV it would more easily stay in a desired position when using DP. Therefore, DP was tested for both cases, resulting in a \textit{low} thruster effect with the extra weight added, and a \textit{high} thruster effect without the extra weight.

\subsubsection{Forwards motion effects} To examine the impact of a sudden forwards movement from the ROV would have on the fish, the vehicle was kept stationary using DP at a depth of approx. 6 meters for 10 minutes. It was then manually controlled in a straight forwards movement lasting approx. 8 seconds, before a new stationary period of 10 minutes was started. The movement was stopped before getting too close to the fish to avoid collisions. 

\subsubsection{Turning effects} To examine the impact of a turning movement while staying a fixed position, i.e. simulating an inspection operation, the vehicle was kept stationary using DP at a depth of approx. 6.5 meters with constant heading. It was then manually controlled to turn on the spot for approx. 180$^\circ$ before a new stationary period with constant heading for 10 minutes was started. 

\subsubsection{Upwards motion effects} To examine the impact of a sudden upwards movement, the vehicle was kept stationary using DP for a period of 10 minutes. These tests were done at two different depths, approx. 6.5 and 8 meters. After 10 minutes, the ROV was moved upwards all the way to the surface. This was tested both using active controls, and passively using  only the positive buoyancy and no thruster power. The ROV was then moved back down, before a new stationary period of 10 minutes was started. 

\subsection{Data Acquisition}
As shown in Figure~\ref{fig:illustraioncases}a, the ROV was equipped with two Ping360 scanning image rotating, single-beam sonars (BlueRobotics Inc.) and a stereo camera~\cite{saad2024stereoyolo+}. The Ping360 sonars were attached facing upwards and downwards, scanning a circular plane right above and right below the ROV with radius of 5 meters. Video data from the stereo camera was saved and used in later analyses: Both quantitative analyses using stereo vision methods, and qualitative blind study analyses of the visible changes in fish behaviour during the different case studies. 

\begin{figure*}[htbp]
 \centering
    \begin{subfigure}[t]{0.32\textwidth}
        \includegraphics[width=1\textwidth]{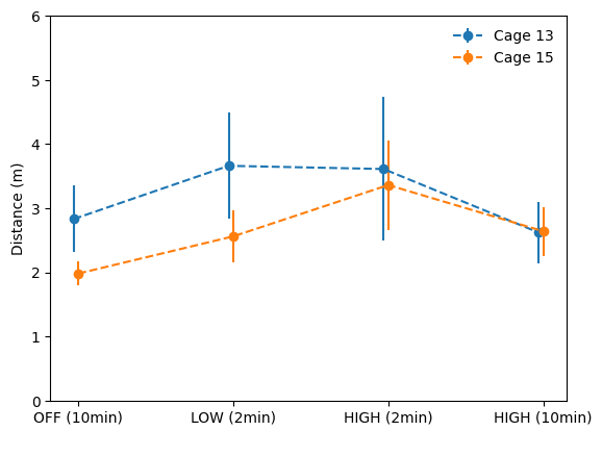}
        \caption{Sonar data DL Method}
        \label{fig:thrustereffectsPing}
    \end{subfigure}
    \begin{subfigure}[t]{0.32\textwidth}
        \includegraphics[width=1\textwidth]{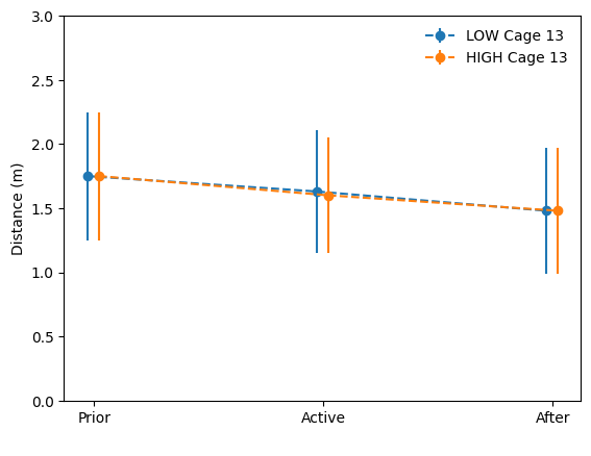}
        \caption{Stereo vision method - Cage 13}
        \label{fig:thrustereffectsSV13}
    \end{subfigure}
    \begin{subfigure}[t]{0.32\textwidth}
        \includegraphics[width=1\textwidth]{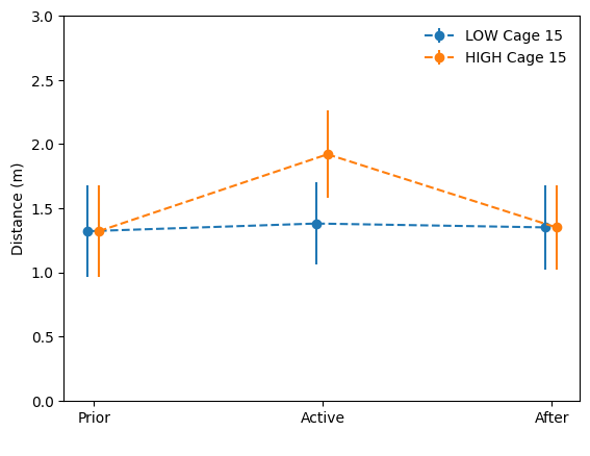}
        \caption{Stereo vision method - Cage 15}
        \label{fig:thrustereffectsSV15}
    \end{subfigure}
\caption{Thruster effects on fish behaviour}
\label{fig:thrustereffects}
\end{figure*}

\section{Methods}
The sonar and stereo camera data collected during field trials were analysed using several different methods, giving both quantitative and qualitative measurements of the changes in fish behaviour. 

\subsection{Sonar data analysis}
A Deep Learning (DL) semantic segmentation method was used in this study, already tested on preliminary experiments in fish farms using a simpler test structure instead of an ROV~\cite{zhang2024farmed}. The method is designed to automatically detect the average distance the fish surrounding the sonar will keep. Because the sonar uses rotating single beam signals, this method was only used to analyse the fish distribution when the ROV kept a stationary position in \textit{position}, i.e. did not have any forwards or upwards movements. 


\subsection{Stereo Vision Method}
Based on prior experiments conducted in~\cite{jakobsen2023deep}, a complete detection, tracking and 3D position estimation pipeline designed to track fishtails in industrial sea cages was proposed in~\cite{alvheim2024identification}. Different methods for the stereo camera system were used to estimate the distance to the fish and the changes in fish behaviour. Because the distance from the fish to the camera is important for our understanding on how comfortable the fish is with the structure introduced to its environment - in this case an ROV - the minimum distance between fish and camera was chosen as one of the metrics for further analysis. The method utilize a combination of stereo vision by SuperGlue, triangulation and RAFT-Stereo, image pre-processing including morphological area opening and discrete wavelet transform, object detection by YOLOv8, and multi-object tracking by ByteTrack to estimate the relative distance between a stereo camera system and detected fishtails. More details on this method can be found in~\cite{alvheim2024identification}.

\subsection{Qualitative Results from video data}
To perform a qualitative blind study of the fish reaction, video close to the beginning of each movement from the ROV was split into 5 second-segments. These videos were then shuffled in a more random order, and viewed by 4 reviewers, giving a numbered score between 5 and -1 for the level of escape response visible in the video, -1 representing attraction. Based on this it was possible to say something about which case studies gave the most significant reaction from the fish. 

\begin{table*}[]
    \centering
\begin{tabularx}{0.8\textwidth} { 
  | >{\raggedright\arraybackslash}X 
  | >{\centering\arraybackslash}X 
    | >{\centering\arraybackslash}X
      | >{\centering\arraybackslash}X
  | >{\centering\arraybackslash}X | }
 \hline
 \textbf{Case study} & \textbf{Sonar deep-learning Method} & \textbf{Stereo Vision Method} & \textbf{Blind Study} & \textbf{Reaction}\\
 \hline
 Thrusters off  & x & x & -- & No\\
 
\hline
Thrusters on low  & x  & x & -- & No\\
\hline
Thruster on high  & x  & x & -- & Fish keep greater distance\\
\hline
Forwards  & --  & x & x & Strong escape response\\
\hline
Turning  & --  & x & x & Weak, fish keep slightly greater distance\\
\hline
Upwards passive  & --  & x & x & Weak increase in distance in one cage\\
\hline
Upwards active  & --  & x & x & Increase in distance in one cage\\
\hline
\end{tabularx}
    \caption{Overview: The different cases were analyzed using three different methods}
    \label{tab:my_label}
\end{table*}

\section{Preliminary results}
This paper present initial findings from the preliminary analysis of the extensive dataset collected from field trials performed in SINTEF ACE facilities. The results discuss the behaviour change of fish in the presence of an ROV. To study the global fish distribution around ROV and the thruster effects, the sonar and video data are utilized. However since Ping360 sonar is getting measurements every 8 sec, only computer vision methods are suited studying the immediate behaviour change of fish during turning, forward and upwards movements. Table \ref{tab:my_label} presents the investigated cases and the conclusions of the study performed in this paper.

\begin{figure*}[htbp]
 \centering
    \begin{subfigure}[t]{0.45\textwidth}
        \includegraphics[width=0.8\textwidth]{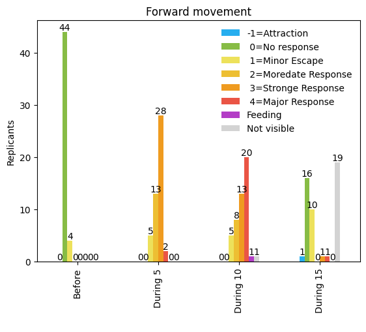}
        \caption{}
        \label{fig:ForwardALL}
    \end{subfigure}
    \begin{subfigure}[t]{0.45\textwidth}
        \includegraphics[width=1.1\textwidth]{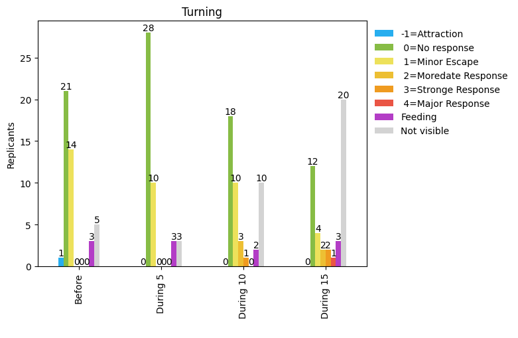}
        \caption{}
        \label{fig:TurningALL}
    \end{subfigure}
    \hspace{1em}\\
    \begin{subfigure}[t]{0.45\textwidth}
        \includegraphics[width=1.1\textwidth]{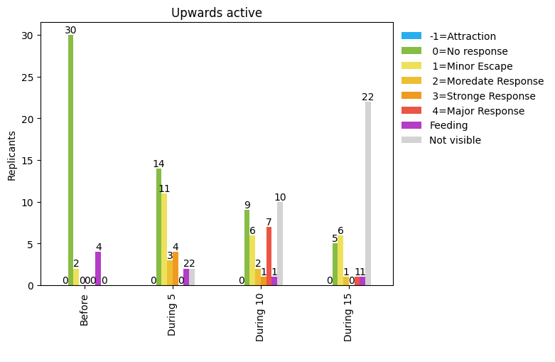}
        \caption{}
        \label{fig:UpwardsActiveALL}
    \end{subfigure}
    \begin{subfigure}[t]{0.45\textwidth}
        \includegraphics[width=1.1\textwidth]{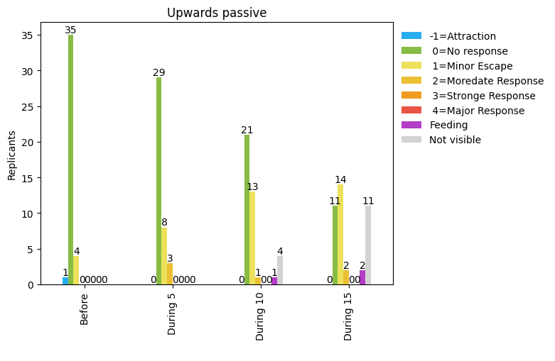}
         \caption{}
        \label{fig:UpwardsPassiveALL}
    \end{subfigure}
    \caption{Results from the blind analysis: Collective Forward, Turning, Upwards Active and Upwards Passive.}
\label{fig:blindanalysiscollective}
\end{figure*}

\subsubsection{Thrusters off - on low - on high}
The sonar data DL method analysis indicated that for one of the two cages there was not much difference in the distance the fish would keep to the ROV when the thrusters were off and on \textit{low} DP, as shown in Figure~\ref{fig:thrustereffectsPing}. However, when the DP and thrusters were on HIGH, the fish would keep a significantly greater distance. Note that this difference could not be seen clearly based on the replicated from the other cage. Visual observation and the number of detected fish from the stereo vision method (Figure~\ref{fig:thrustereffectsSV13} and~\ref{fig:thrustereffectsSV15}) indicate that the reason for this is mainly the very noisy video data obtained in cage 13.

\subsubsection{Forwards movement}
The stereo vision method analysis also showed that the fish would 'move closer' to the ROV during the forwards movement when the ROV drove towards the fish, but as the stereo camera only gives us the distance to the fish in the direction of the movement, this only tells us that the ROV drove towards the fish, and not much about the general escape response. The preliminary analysis of the blind study also shows that the strongest reaction from the fish could be seen when the ROV had a sudden forwards movement (Figure~\ref{fig:ForwardALL}), where all reviewers from the blind study agreeing that there was a strong reaction from the fish in every case, with all the fish eventually disappearing from the camera view.

\subsubsection{Turning movement} 
For the blind study analysis of the turning movements, the majority of the fish did not seem to have any reaction for most cases, though there were a few stronger escape responses (Figure~\ref{fig:TurningALL}). However, the analyses using the stereo vision method showed that for the turning movements, the fish would in most cases keep a greater distance to the ROV during the turning movements, and then come closer again once the turning movement was over, though there was also cases where the fish would move closer also during the turning movement. Further, analysis is required to investigate if the correlation between swimming direction of fish and ROV.

\subsubsection{Upwards movement - active and passive}
The analysis using the stereo vision method showed that for cage 13, the fish would keep a significantly larger distance to the robot during active upwards movements using the thrusters, while in cage 15 there was no clear difference. For cage 13 there was a slight increase in the distance the fish kept during the passive upwards movement, though much less significant, while for cage 15 there was no difference for the passive movement. 

For the blind study, there was also also possible to determine some reaction from the fish during the upwards movements, and there seemed to be a significant difference between the passive floating movement and the active movement using thrusters: While parts of the fish did not seem to react to any of the upwards movement, the fish that \textit{did} react had a much stronger escape response for the thruster driven upward movements than for the passive movements, as seen in Figure~\ref{fig:UpwardsActiveALL} and Figure~\ref{fig:UpwardsPassiveALL}.

\section{Conclusions and future work}
The preliminary results presented in this paper indicate that different factors might affect the fish behaviour in industrial scale fish farms. The analyses show that the thrusters effects, forwards and active upwards movements seemed to have the greatest impact on the fish. For the other more subtle test cases it was not possible to definitely determine  that a specific movement gave strong reactions from the fish. However, the three different methods for analysis support each other and give similar results in cases with strong changes in fish behaviour. These methods should therefore be tested on further field trials to increase the available dataset. Using these for identification and quantification of the impact robot operations have on the fish behaviour, valuable insights can be gained, which will help in the work towards developing dedicated and fish-friendly robotic solutions for the aquaculture environment.





\section*{ACKNOWLEDGMENT}

The work presented in this publication is supported by the Research Council of Norway (RCN) project CHANGE (RCN project no. 313737). The authors would
like to thank SINTEF ACE and Mats Aarsland Mulelid for their support within this project.


\bibliographystyle{plain}
\bibliography{bibliography}

\end{document}